# Optimizing Image Capture for Computer Vision-Powered Taxonomic Identification and Trait Recognition of Biodiversity Specimens

**Authors:**
Alyson East[1#], Elizabeth G. Campolongo[2], Luke Meyers[3], S M Rayeed[4], Samuel Stevens[2], Iuliia Zarubiieva[5,6], Isadora E. Fluck[7], Jennifer C. Girón[8], Maximiliane Jousse[9], Scott Lowe[6], Kayla I Perry[2], Isabelle Betancourt[10], Noah Charney[1], Evan Donoso[11], Nathan Fox[12], Kim J. Landsbergen[13], Ekaterina Nepovinnykh[4], Michelle Ramirez[2], Parkash Singh[2], Khum Thapa-Magar[14], Matthew Thompson[2], Evan Waite[10], Tanya Berger-Wolf[2], Hilmar Lapp[15], Paula Mabee[16], Charles Stewart[4], Graham Taylor[5,6], Sydne Record[1]

**Institutions:**

1. University of Maine, 5755 Nutting Hall, Orono, ME, USA 04469
2. The Ohio State University, 2015 Neil Avenue, Columbus OH 43210
3. University of Puerto Rico Rio Piedras, 6, 2526, 601 Av. Universidad, San Juan, 00925, Puerto Rico
4. Rensselaer Polytechnic Institute, 110 8th St, Troy NY 12180
5. University of Guelph, 50 Stone Rd E, Guelph, ON N1G 2W1, Canada
6. Vector Institute, 108 College St W1140, Toronto, ON M5G 0C6, Canada
7. University of Florida, 1478 Union Rd, Gainesville, FL 32603
8. Texas Tech University, 3301 4th Street, Lubbock, TX 79415
9. McGill University, 1205 Dr Penfield Ave, Montreal, Quebec H3A 1B1
10. Arizona State University, 734 W. Alameda Dr. Tempe AZ 85282
11. National Ecological Observatory Network, 60 Nowelo Street Hilo, Hawaii 96720
12. University of Michigan, 500 S State St, Ann Arbor, MI 48109
13. Antioch College, 1 Morgan Place, Yellow Springs OH 45387
14. University of Colorado, 552 UCB, Boulder, CO 80309
15. Duke University, 2080 Duke University Road, Durham, NC 27708
16. National Ecological Observatory Network, 1685 38th St., Suite 100, Boulder, CO 80301

# Corresponding Author: alyson.east@maine.edu, 5755 Nutting Hall, Room 210, Orono, Maine 04469

# Abstract

1) Biological collections house millions of specimens with digital images increasingly available through open-access platforms. However, most imaging protocols were developed for human interpretation without considering automated analysis requirements. As computer vision applications revolutionize taxonomic identification and trait extraction, a critical gap exists between current digitization practices and computational analysis needs. This review provides the first comprehensive practical framework for optimizing biological specimen imaging for computer vision applications.
2) Through interdisciplinary collaboration between taxonomists, collection managers, ecologists, and computer scientists, we synthesized evidence-based recommendations addressing fundamental computer vision concepts and practical imaging considerations. We provide immediately actionable implementation guidance while identifying critical areas requiring community standards development.
3) Our framework encompasses ten interconnected considerations for optimizing image capture for computer vision-powered taxonomic identification and trait extraction. We translate these into practical implementation checklists, equipment selection guidelines, and a roadmap for community standards development including filename conventions, pixel density requirements, and cross-institutional protocols.

4) By bridging biological and computational disciplines, this approach unlocks automated analysis potential for millions of existing specimens and guides future digitization efforts toward unprecedented analytical capabilities.

## 1. Introduction

Biological collections house millions of specimens, documenting life on Earth. These specimens contain a wealth of biological information (e.g., traits), yet such data and subsequent discoveries from them remain locked away by practical constraints of time, personnel, and funding. However, as digitization efforts advance—with collection information transcribed from physical labels and specimens increasingly captured as digital images (Beaman & Cellinese, 2012; Blagoderov et al., 2012; Eckert et al., 2024; Nelson et al., 2012) and linked back to the physical specimens (Lendemer et al., 2020)—new possibilities emerge. Sharing digitized 'extended specimen' information, including images, openly through biodiversity data platforms (Gallagher et al., 2020; Global Biodiversity Information Facility, 2025; *iDigBio*, 2025; Parr et al., 2014) greatly increases specimen use and impact as digital collections allow widespread access and analysis of high-quality, ecologically important information at unprecedented spatial and temporal scales (Davis, 2023).

Initial digitization efforts aimed to increase data accessibility, create digital backups of physical specimens, produce comprehensive photo-vouchers (Gómez-Bellver et al., 2019), and highlight key taxonomic features for identification and publication. As a consequence, most imaging protocols have prioritized accurate phenotypic representation without considering how these images might be used for computational data extraction, which is now feasible given advances in computational efficiency and computer vision (CV) models (particularly convolutional neural networks, vision transformers, self-supervised learning methods, and diffusion models) (Buckner et al., 2021; Greene et al., 2023; Pollock et al., 2025a; Soltis, 2017). However, computer vision systems process images differently than human observers (Geirhos et al., 2022), requiring specific considerations in image acquisition that may not be incorporated into current protocols. Thus, a gap exists between current imaging practices and the potential for automated analysis.

The intersection of high-quality digitized specimens and modern computer vision creates transformative research opportunities. Computer vision algorithms can drastically reduce the time required to perform tasks like text extraction, morphological trait quantification, and species detection and classification from images (Pepper et al., 2021; Soltis, 2017) while potentially detecting subtle details in shape and color that might be overlooked by the human eye (Hoyal Cuthill et al., 2024; Stoddard et al., 2014). These applications span diverse tasks including automated species identification, morphometric analysis, phenological monitoring, ecological surveys, and biodiversity assessments (reviewed by T.T. Høye et al., 2021; Lürig et al,. 2021). These new technologies have the potential to enhance the accuracy and efficiency of collecting morphological trait measurements (Soltis, 2017), aid in the process of species identification across the tree of life (Hussein et al., 2022; Stevens et al., 2024), and allow for innovative data collection and analysis on a global scale (Pollock et al., 2025b). These emerging computational approaches hold tremendous promise, yet realizing their full potential requires deliberate coordination between biological and computational disciplines.

Integrating biological collections into CV workflows represents an emerging interdisciplinary frontier with unique challenges. These fields have developed distinct methodologies, technical languages, and research priorities that can create unintentional barriers to collaboration. When biological specimens are imaged without consideration for computational analysis, and when CV systems are developed without awareness of the complexity of biological variation, research effectiveness suffers. Because of this, valuable existing datasets may remain underutilized while resources are directed toward generating new data. Addressing this interdisciplinary gap through standardized specimen image acquisition protocols based on findability, accessibility, interoperability, and reusability (FAIR) principles (Wilkinson et al., 2016) is fundamental to advancing research in taxonomy, phenomics, ecology, and evolution (Houle et al., 2010; Lürig et al., 2021).

Given this interdisciplinary gap, we brought together a diverse group of researchers from various disciplinary backgrounds and career stages at two venues hosted by the Imageomics Institute at The Ohio State University to determine CV context for imaging biological specimens and to synthesize emerging techniques to optimize museum specimen imaging protocols for integration with artificial intelligence (AI) systems. The first venue, Beetlepalooza, was a workshop that brought together taxonomists, collection managers, ecologists and computer scientists to tackle challenges in automated analysis of entomological specimens (Lapp et al., 2024). The second venue was a graduate-level “Experiential Introduction to AI and Ecology” course where students and faculty from biology and computer science collaborated on practical solutions for integrating computer vision with biological collections. These interactions generated a comprehensive synthesis of interdisciplinary knowledge by addressing evidence-based guidance from real-world case studies that bridges disciplines by consolidating imaging standards for biologists and AI developers.

## 2. Computer Vision Context for Imaging Biological Specimens

### 2.1 Model Generalization

In taxonomic identification, biologists often refer to guides containing representative specimens that display typical characteristics of a species. However, taxonomists recognize that real-world specimens frequently deviate from these idealized examples due to developmental stages, physical condition, and sexual or phenotypic variation. Through extensive exposure to such variability in morphologies, taxonomists develop the ability to identify species despite these differences. Similarly, CV systems must be designed to handle the gap between controlled training conditions and real-world application scenarios.

Modern CV approaches offer several strategies to build robust models, such as data augmentation (Shorten & Khoshgoftaar, 2019; Yun et al., 2019; Zhang et al., 2018), domain adaptation methods (Tzeng et al., 2017; Wang & Deng, 2018), or transfer learning from foundation models trained on diverse data (Kolesnikov et al., 2020; Kornblith et al., 2019; Yosinski et al., 2014). However, these technical solutions all rely on one fundamental principle: models need exposure to variation in training data to learn how to robustly handle such variation during future use. Consider a model trained to measure butterfly wing size. If trained only on perfectly spread specimens photographed from above, the model may fail when encountering folded wings or angled views. However, with appropriate training data spanning these variations, models can learn to extract meaningful features despite such differences.

**2.2 Data Leakage**

Data leakage is a term used for a family of issues that can befall CV projects that make models difficult to verify or ineffective. Consider a taxonomic identification model trained to distinguish mosquito species, or a trait extraction model designed to measure butterfly wing size. For such applications, data leakage occurs in two cases: (1) inadequate separation of training, validation, and testing data, and (2) when models learn to utilize non-generalizable image features to achieve their target task (Kapoor & Narayanan, 2023). Both cases lead to an overestimation of model accuracy and hinder model generalization.

Proper separation of training, validation, and test data relies heavily on accurate tracking of unique specimen identifiers, as discussed in Section 3.1 — without these identifiers, researchers risk inadvertently including the same specimen in both training (or validation) and test subsets, artificially inflating performance metrics (Fig. 1a).

When designing imaging plans for biological specimens, it is helpful to anticipate generalization and data leakage challenges that may occur when used for AI/ML applications. For instance, preservation methods can alter specimen appearance through color changes or physical deformation. Digitization methods may introduce differences (e.g., in resolution, lighting, and background elements [Fig. 1b]) and confounding factors. For model generalization and effective use, sufficient representative samples of images illustrating variation must be included during model training, or the source of variation should be eliminated using imaging standardization techniques. This fundamental tension between standardization and variation underlies many of the practical considerations in subsequent sections.

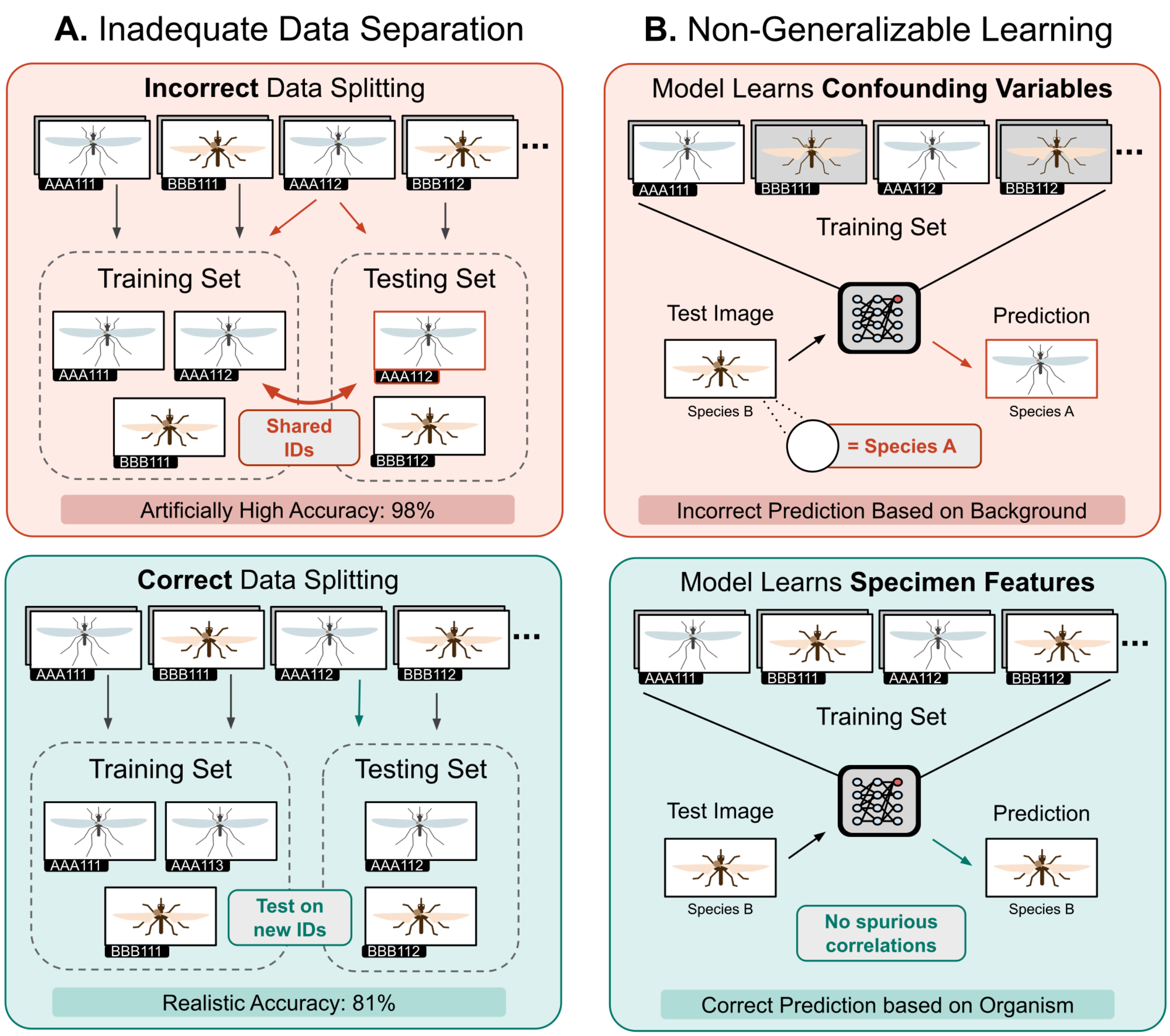

Figure 1: The example above highlights data leakage in a hypothetical classification model trained to distinguish mosquito species from images. Data leakage occurs through (A) inadequate separation of training and test data, where the same specimen (AAA112) appears in both datasets leading to artificially inflated performance metrics, and (B) models learning non-generalizable features such as backgrounds or imaging artifacts rather than biologically relevant specimen

characteristics. Proper specimen tracking and standardized imaging protocols prevent both types of leakage.

We emphasize data leakage and model generalization upfront, as they inform the scope and reasoning behind the following suggestions and highlight the importance of considering future model use when planning and executing image acquisition protocols. When following the recommendations outlined in subsequent sections, it is essential to focus on capturing images that enable models to learn reliable information about the specimens being examined. Part of this process also involves detailed documentation of the factors that could cause data leakage in a comprehensive metadata trail, to be able to detect and address any issues during future model development.

## 3. Considerations for Imaging Biological Specimens

Recommendations from our interdisciplinary group fall into two categories: optimized image capture, data quality and usability (Fig. 2). These interconnected considerations form a framework for the successful integration of biological specimen images into CV pipelines, enabling researchers to maximize the scientific value of both historical and newly collected specimens.

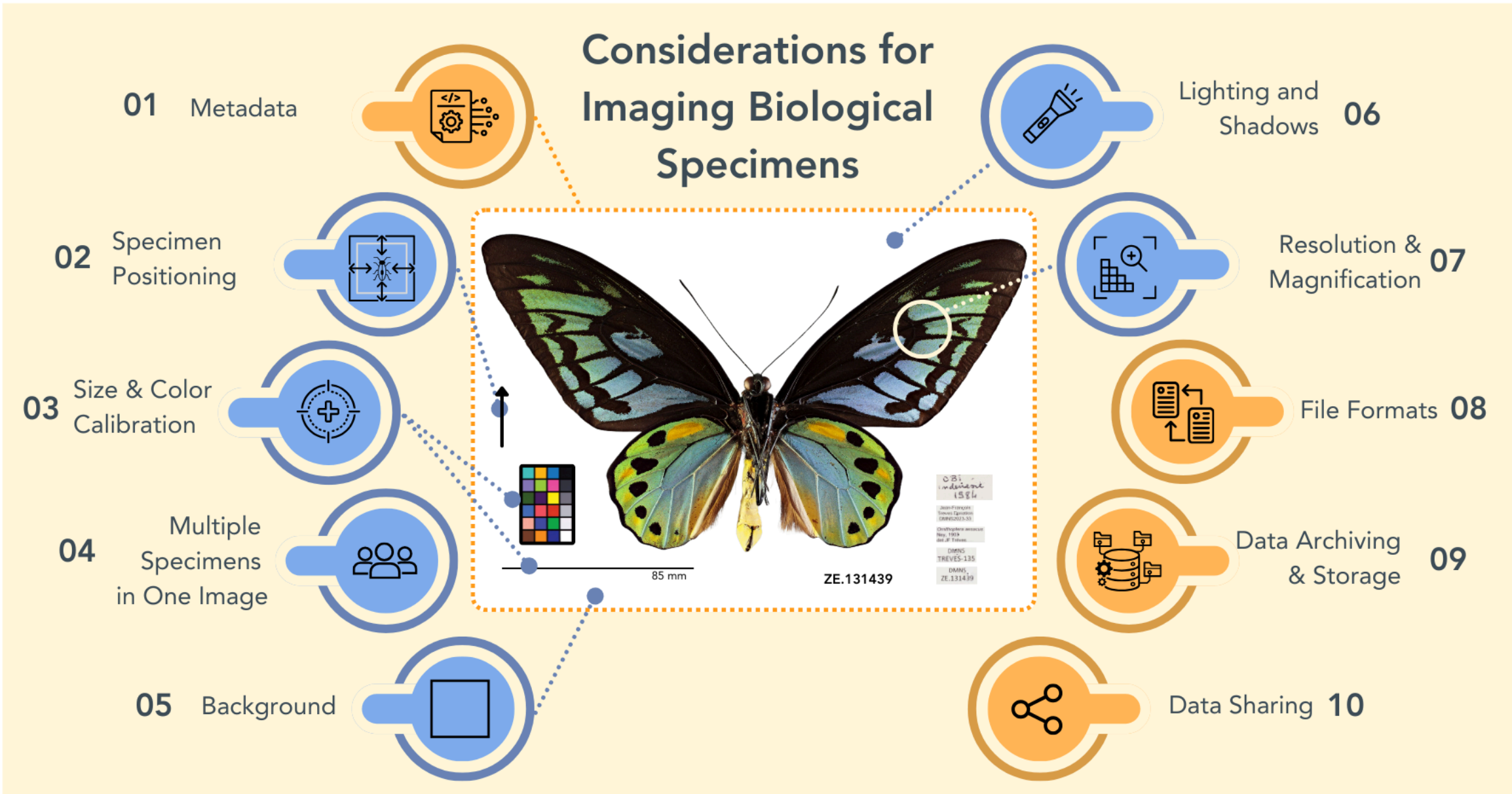

Figure 2: Overview of the considerations for creating biological specimen images suited for computer vision tasks. Topics are grouped into two functional categories by color: blue for optimizing image capture, and orange for data quality and usability. *Ornithoptera aesacus* (Ney, 1903) butterfly image adapted from (Denver Museum of Nature & Science - Entomology, ZE.131439) under CC BY-NC from DMNS (2024).

### 3.1 Metadata

Image collection protocols should have a comprehensive plan for the collection of metadata in addition to the images themselves (e.g., information on the images, protocols for taking photographs, and specimens). Well-maintained metadata provides biological context for model development and reproducibility in an efficient format, bridging the documentation needs of biological collections and CV applications (Barman & Martini, 2020). In biological applications where sample sizes are often constrained, thorough metadata serves as a safeguard against data loss by enabling researchers to trace and correct errors in specimen handling, identification, or data transcription (Karnani et al., 2024).

We emphasize the importance of thorough documentation of metadata for an individual image, including specimen and imaging information (Table 1). Specimen metadata links each image to the physical specimen(s) being photographed. The most critical element is the unique identifier for each specimen, which serves as the foundation for all downstream analyses and data integration. These identifiers must be reliably recorded alongside image filenames in structured metadata files (e.g., CSV, JSON) and, whenever possible, included within the image itself for redundancy (e.g., to be read by a computer via optical character recognition [OCR] and cropped before analysis). The relationship between images and specimen data is best maintained through relational approaches: each image filename connects to a metadata record containing the specimen identifier(s), taxonomic information, collection details, and imaging parameters. While an informative, universal file naming convention is desirable, this approach accommodates the reality that different institutions use different identifier systems and naming conventions while ensuring that computational results can always be linked back to biological context. Without maintaining this structured connection between images and specimen identifiers, computational results cannot be traced to their biological origins, effectively severing the ties between trained models and the broader scientific value of the underlying collections.

Collections often contain varying levels of specimen detail,  with rare specimens receiving comprehensive records while common species may be housed in bulk without individual records. Even bulk specimens require a batch-level identifier at the finest available resolution, with metadata clearly documenting traceability limitations. Additional biological information (e.g., taxonomic identification, life stage, sex, or reproductive state) and collection data (e.g., geographic information, collection date) should be preserved to enable domain-relevant analyses and integration with existing biodiversity databases.

Imaging metadata links images to the settings and protocol used to photograph them (Table 1). It may include information that is stored in the image file as Exchangeable Image File Format (EXIF) data (e.g., camera settings, filenames), but should also include details that are not stored in the image file. For example, metadata describing the imaging protocol (e.g., camera and lens types, lighting, use of diffusion paper, tripod mounting heights) and acquisition (e.g., who, where, and when the image was taken). It is important to record such information that may not be immediately extractable from the images themselves to provide context needed to understand model generalization.

Table 1: Types of Metadata with their collection purpose and notes on methodology. QC refers to Quality Control, QA to Quality Assurance, ODA to Object Detection Algorithm, and OCR to Optical Character Recognition. **Additional specimen metadata may also include information about the life stage of the specimen (e.g., juvenile or adult), quality of the specimen (e.g., detached wings, full body), or potentially less-discernible characteristics (e.g., reproductive state). Where possible, including biologically-meaningful metadata allows for more domain-relevant AI.

| **Metadata Category** | **Data Type** | **Purpose** | **Machine Extractable** |
|---|---|---|---|
| **Specimen** | Unique Identifiers (individual ID or batch ID) | Error tracking, QA, data linkage | Potentially (OCR) |
| | Sample Origin (ex. field plot location) | Error tracking, QA, data linkage | Potentially (OCR) |
| | Geographic Location of Collection | Error tracking, QA | Potentially (EXIF or OCR) |
| | Collection Date | Error tracking, QA | Potentially (OCR) |
| | Taxonomic Information | Classification verification, model training | Potentially (OCR) |
| | Additional Biological Data** | Classification verification, model training | No |
| **Imaging** | Resolution | QC, reproducibility, size calibration | Yes |
| | Timestamp | Error tracking, QA | Yes (check at setup) |
| | Exposure (shutter, aperture, ISO, flash) | Image standardization | Yes |
| | White Balance | Image standardization | Yes |
| | File Format | Processing compatibility | Yes |
| | Scale Bar Locations | Size & color calibration | Yes (w/ supervision - ODA) |
| | Camera Specifications | Image standardization | No |
| | Lens Specifications | Image standardization | No |
| | Flash Specifications | Image standardization | No |
| | Subject Orientation (e.g., ventral, dorsal) | Processing verification, model training | No |
| | Subject morphology (e.g., head, leg) | Processing verification, model training | No |
| | Geographic Location of Imaging | Error tracking, QA | Sometimes (check at setup) |
| | Photographer | Error tracking, QA | No |
| | Date Photographed | Error tracking, QA | Yes (check at setup) |

For efficient metadata management at scale, we recommend utilizing established standards for biological data (Baskauf et al., 2023; Darwin Core Maintenance Group, 2023; Morris et al., 2013), which provide a generalized framework for digitizing collections. Metadata recorded during imaging should be organized in a form that can be easily exported to machine-readable formats (e.g., CSV, JSON, COCO) . Auxiliary information, such as metadata category descriptions, should be defined/described in an associated, but separate data file or document to provide context for those unfamiliar with the dataset. Though some of the information may seem intuitive or understood within a discipline, it may not be as clear across disciplines or may simply get lost at scale — for AI training, a human will not examine each individual image to ascertain this information. While diligent recording of metadata can be time-consuming with large volumes of data, additional efficiencies in metadata curation are possible utilizing CV methods such as Object Detection and Optical Character Recognition algorithms to automate data recording that would otherwise require manual transcription (Lürig et al., 2021; Tolstaya et al., 2010).

Metadata should be treated as essential data products requiring careful planning. Though time-consuming to record, detailed metadata enables both image analysis projects and broader biological studies. Each imaging decision outlined in subsequent sections represents a potential component that should be documented in metadata.

**3.2 Specimen Positioning**

Specimen positioning is important in two contexts: orientation of the specimen and location in the camera focal frame. The orientation of the specimen may affect model performance and needs to be appropriate for the downstream application. For trait segmentation applications, establish standardized specimen orientations and consistent appendage positioning for comparative morphology. Use of positioning guides or templates to ensure reproducibility across specimens and imaging sessions increases success of detection and segmentation (Z. Feng et al., 2025; Ravi et al., 2024). By contrast, for classification models trained to identify species, including a diverse sample of images showing various specimen orientations would improve model generalization.

Another consideration in specimen positioning is that the location of a specimen and a reference scale (see Section 3.3) in a focal frame matters when imaging with a camera as opposed to a scanner. Distortion is an inherent optical artifact in camera lenses that affects the accuracy of size, particularly around the periphery of an image. In biological imaging, these distortions can introduce inaccuracies in trait measurement and representation of specimen shape, which then affect downstream uses.

Distortion depends upon the lens design and typically manifests in two primary forms: barrel distortion, which is more common in lenses with shorter focal lengths, and pincushion distortion, seen in lenses with longer focal lengths (Fig. 3). For projects requiring accurate measurements, systematic lens characterization is essential. Test specific lens by photographing a calibration grid to identify the zone of distortion (OpenCV Docs, 2024). Document calibration results and distortion maps in metadata to enable consistent specimen positioning and measurement correction. Position specimens and scale bars within the identified minimal distortion zone to ensure measurement accuracy across different equipment setups. Avoid placement of specimens and scale or color bars near edges or corners of the image frame where distortion effects are greatest (Fig. 3).

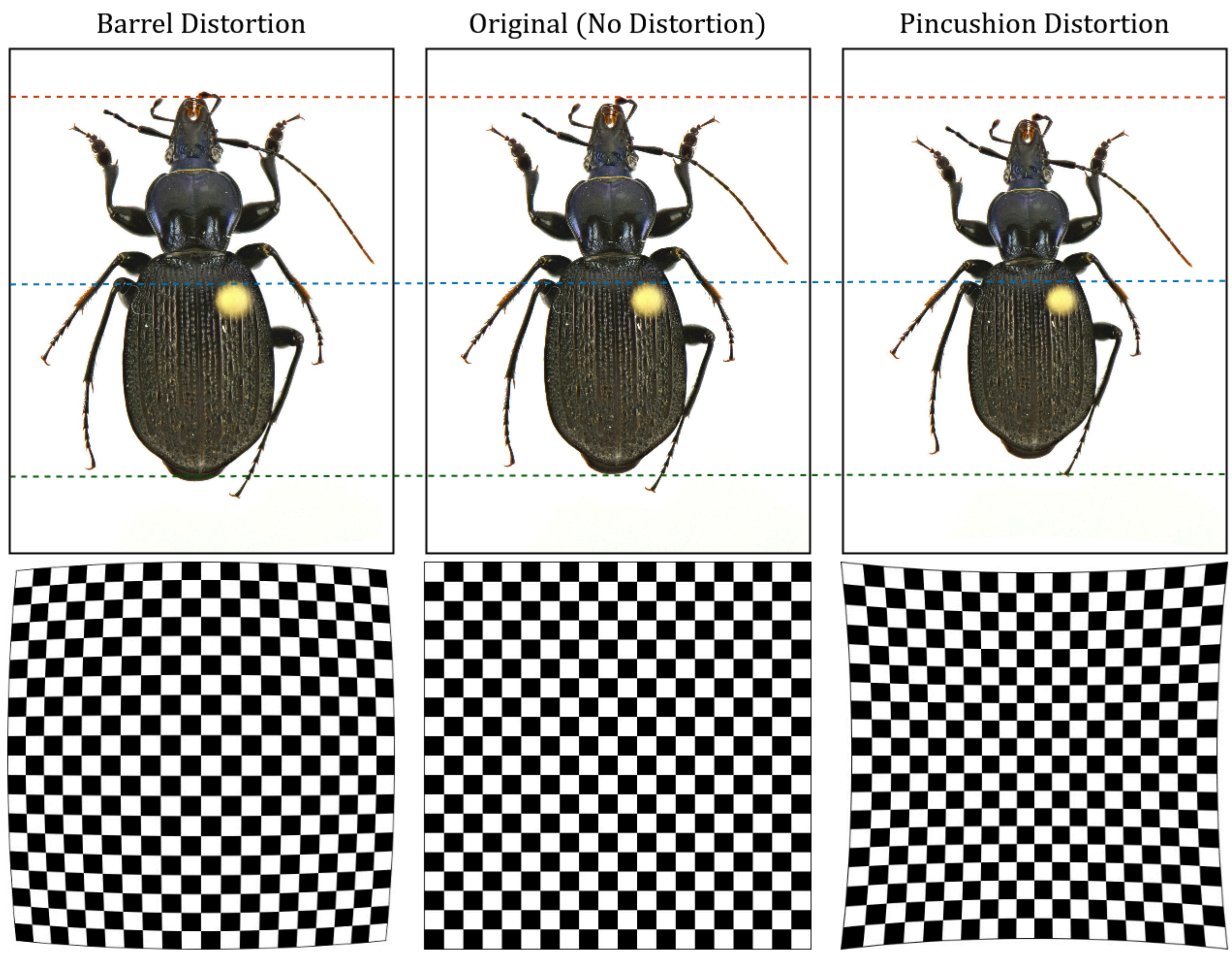

Figure 3: Examples of types of lens distortion and the effects on pixel size appearance across an image (top row) and a checkerboard (bottom row). The top row demonstrates the effect of distortion on a manipulated image of a *Sphaeroderus bicarinatus* (LeConte, 1853) specimen with guiding lines to demonstrate the subtle changes. The bottom row shows the distortion clearly on a checkerboard field with the original undistorted board in the middle, barrel distortion (shorter focal lengths) to the left, and pincushion distortion (longer focal lengths) to the right. Image credit: Essig Museum of Entomology (EMEC-EMEC) via National Ecological Observation

Network Biorepository at Arizona State University. Accessed via NEON (National Ecological Observatory Network) Biorepository data portal 07-03-2025. Sphaeroderus bicarinatus EMEC-EMEC: EMEC1311627. Photograph.
https://biorepo.neonscience.org/imglib/neon/EMEC_EMEC/NEONBET/SPHBIC-Dorsal.jpg

**3.3 Size and Color Calibration**

For reproducibility and size measurements, physical scales should be included in every photo. While some traits can be scale agnostic (e.g., presence-absence variables, textures), a reference scale is requisite for the recovery of many meaningful trait measurements (e.g., size, area). Consistent placement of scale bars facilitates automated removal during image processing, and this placement should be outside of the zone of distortion (Section 3.2). For size measurements, it is important to include a scale bar in the same focal plane as the specimen to establish accurate pixel-to-length ratios, and that the camera be oriented level with the subject (Fig. 4). Positioning the reference scale and subject in the same plane minimizes distortion errors and enables reliable measurements across different images. Maintaining a fixed camera-to-specimen distance with a consistent setup creates predictable scaling relationships using pixel-based measurements, but sacrifices the ease with which measurements can be connected across datasets. In CV applications where image-derived size measurements are calculated, there is also a need for validation against physical measurements using precision instruments (e.g., calipers) to ensure accuracy.

Color accuracy can be crucial for specimen documentation and CV applications, as color features often serve as key diagnostic characteristics for species identification and condition assessment (Hussein et al., 2022; Sunoj et al., 2018). Including a standardized color reference chart in each image or maintaining consistent lighting and background conditions can improve color fidelity (Fig. 4). Color charts enable manual and automated correction during processing while providing calibration points for CV models. Additionally, color calibration scales can serve to standardize across institutions and datasets.

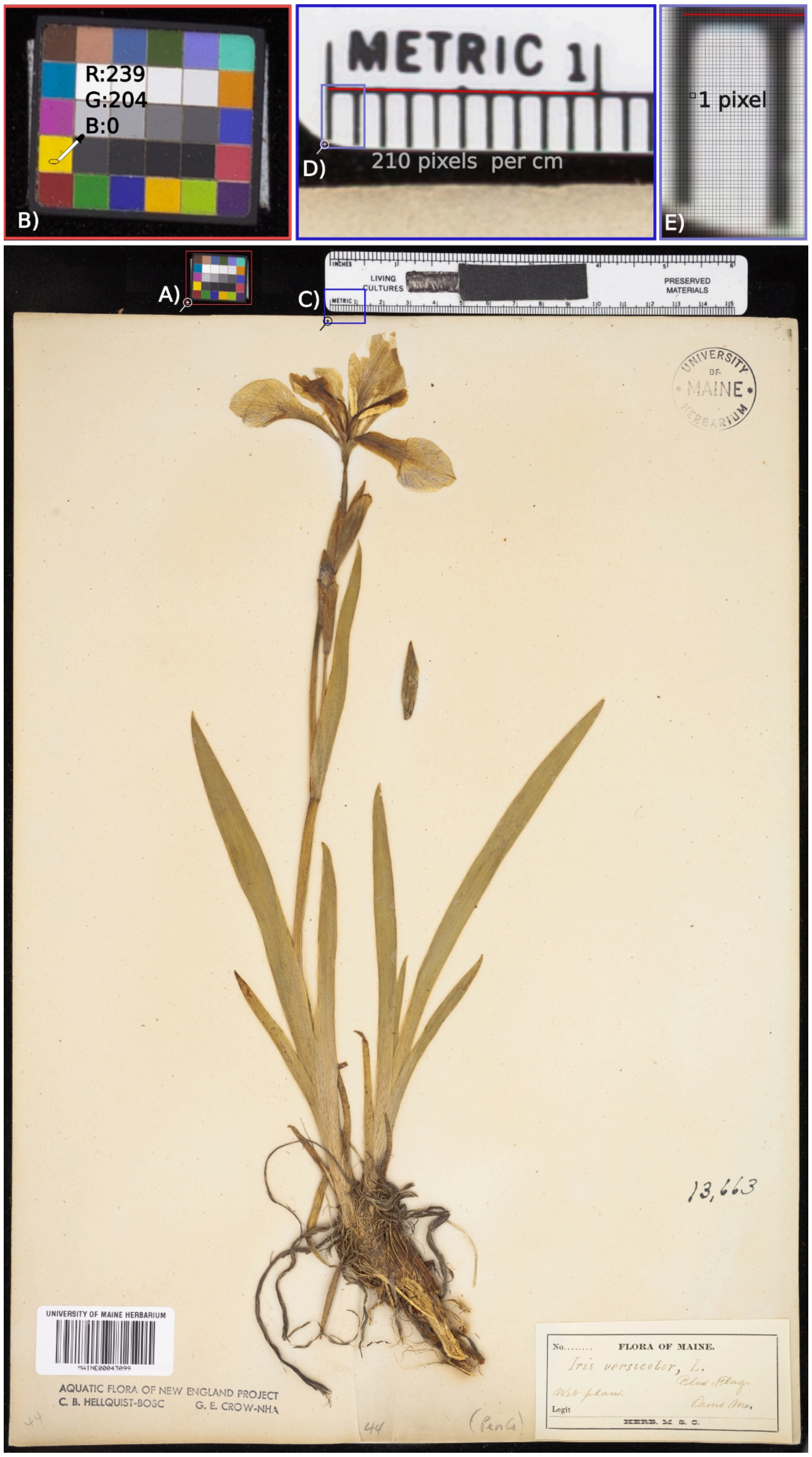

Figure 4: Example of an *Iris versicolor* L. (determined by Hellquist & Crow) herbarium specimen. A) color palette scale bar, B) magnified color palette with associated RGB color reading for calibration, C) measurement scale bar, D) single centimeter magnified with annotation of visual representation of pixel-based length calibration, E) resolution at which individual pixels become

visible for reference of pixel-based length measurement. Image modified from Herbarium specimen data provided by: University of Maine (Accessed through the Consortium of Northeastern Herbaria website, www.neherbaria.org, 2025-03-03. Occurrence ID: urn:uuid:5e4cc6a7-5c1b-45d3-8873-9ca399264e0b).

### 3.4 Multiple Specimens

Some collections may require imaging multiple specimens at once. While single-specimen imaging is ideal when feasible, many existing collections require multiple specimens per image due to time constraints on imaging large collections, and where existing historical practices (e.g., storing pinned insect specimens close together in trays for space efficiency) and specimen fragility are of particular consideration. The use of images containing multiple specimens can be leveraged by implementing adjustments to capture and metadata tracking protocols. If multiple specimens are in the same image and individual unique identifiers are available, object detection (Liu et al., 2024) and cropping can be used to generate a bounding box around the individuals. Images cropped to those bounding boxes output individual images that can be referenced back to the original image and all accompanying metadata. It is imperative to maintain a consistent naming convention for cropped images of individuals with clear linkages back to metadata tied to the original image. The separation of individuals is essential for accurate trait segmentation, as overlapping individuals would yield erroneous measurements.

An additional complication is introduced if the specimens contained in a single image are of different taxa because differently-sized taxa may require different imaging techniques and morphological similarities among taxa introduce the potential for error in downstream classification models. When imaging taxa of different sizes, it is important to quality-check the clarity and resolution of individuals in the image to ensure that the lens and settings used capture information at a high enough resolution to resolve details on smaller-sized taxa. Additionally, it is important to ensure that there is no physical overlap of specimens. Even though (re)arranging individuals to maintain physical separation can represent a serious manual effort, it increases the performance of object detection, eases image data processing, and removes potentially confounding information for training species identification models (Fig. 5).

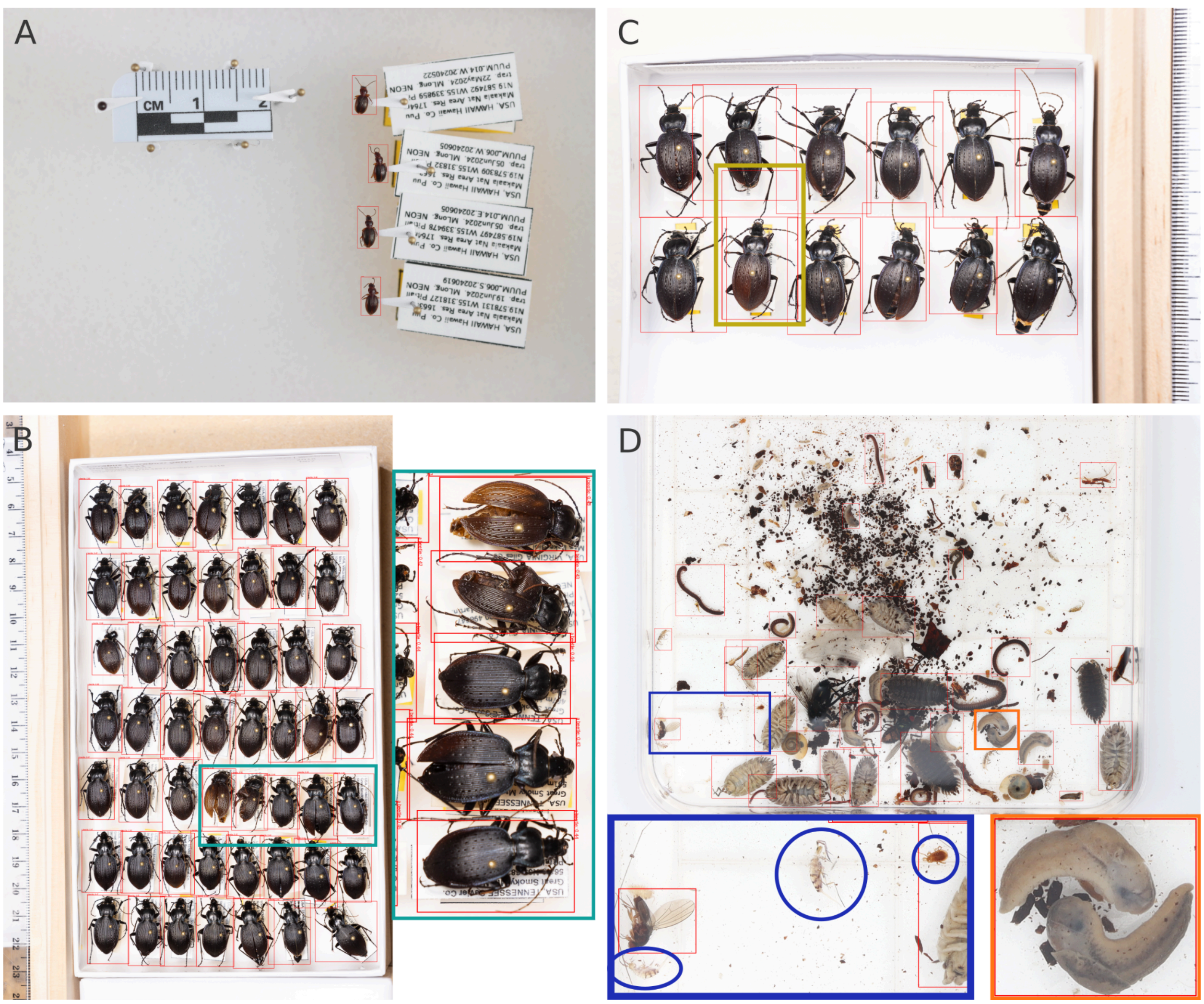

Figure 5: Object detection challenges with multiple specimens. Red boxes show Grounding DINO object detection (Liu et al., 2024) results where the algorithm recognizes individual specimens. (A) Ideal scenario: individual *Mecyclothorax rufipennis* Liebherr, 2008 in separate bounding boxes (East 2025a). (B) Partial success: all individual *Carabus goryi* Dejean, 1831 are detected, but bounding boxes include neighboring specimens (teal box). (C) Detection confusion: overlapping specimens of *Carabus goryi* Dejean, 1831, with one individual detected twice in bounding boxes of different sizes (yellow box). (D) Complex scenario of pitfall trap bycatch (Thibault et al., 2023) showing multiple detection failures (blue circles indicate missed specimens) and errors (orange box shows two specimens detected as one). Image sources: A) Alyson East, University of Maine, Funding: S. Record (NSF 242918). Accessed via NEON (National Ecological Observatory Network) Biorepository data portal 01-19-2025. Puʻu Makaʻala *Mecyclothorax Rufipennis*. Photograph. B) I.S. Betancourt, National Ecological Observation Network Biorepository at Arizona State University. Accessed via NEON (National Ecological Observatory Network) Biorepository data portal 20240808-Carabus_goryi_d07-7.jpg. Photograph. flickr.com/photos/neonsciencedata/54498665352/in/album-72177720319044964 C) I.S. Betancourt, National Ecological Observation Network Biorepository at Arizona State University. Accessed via NEON (National Ecological Observatory Network) Biorepository data portal 01-19-2025. 20240807-Carabus_goryi_NEON_BET-D01-5.jpg. Photograph. flickr.com/photos/neonsciencedata/54499793468/in/album-72177720319044964 D) Alyson East, University of Maine, Funding: S. Record (NSF 242918). Accessed via NEON (National Ecological Observatory Network) Biorepository data portal 01-23-2025. Puʻu Makaʻala Bycatch. Photograph.

**3.5 Background**

In CV, particularly when employing neural networks, the relationship between a specimen and its background significantly impacts model performance in identification and classification applications (Xiao et al., 2020). When backgrounds are inconsistent—varying in color, texture, or content—they can introduce distractions that interfere with the model's ability to identify relevant features of the specimen itself. Backgrounds also influence visual perception and color representation. Selecting a consistent, non-reflective, neutral, and uniform background can substantially reduce unwanted variations in specimen appearance. This consistency allows models to better isolate and learn the intrinsic features of the specimens rather than incidental background characteristics.

To demonstrate how image backgrounds influence the ability of CV models to focus on the individual of interest, we analyzed images using attention visualization techniques on the species classification model BioCLIP  (Stevens et al., 2024). In this classification task, the model's objective was to correctly identify the specimen in the image as *Phyllobates terribilis*. Attention mechanisms in neural networks highlight regions of an image that most strongly influence the model's predictions. When visualized using techniques like Gradient-weighted Class Activation Mapping (Grad-CAM) (Selvaraju et al., 2020), it is possible to see a heat map indicating what information on the image most informs the model inferences. Our example shows that when backgrounds are removed from specimen images, attention maps become more concentrated on the specimen itself, as opposed to the image background (Fig. 6). The effect of image background has important implications for model training, as background elements may either provide useful contextual information (as in field photographs) or create misleading associations with collection conditions.

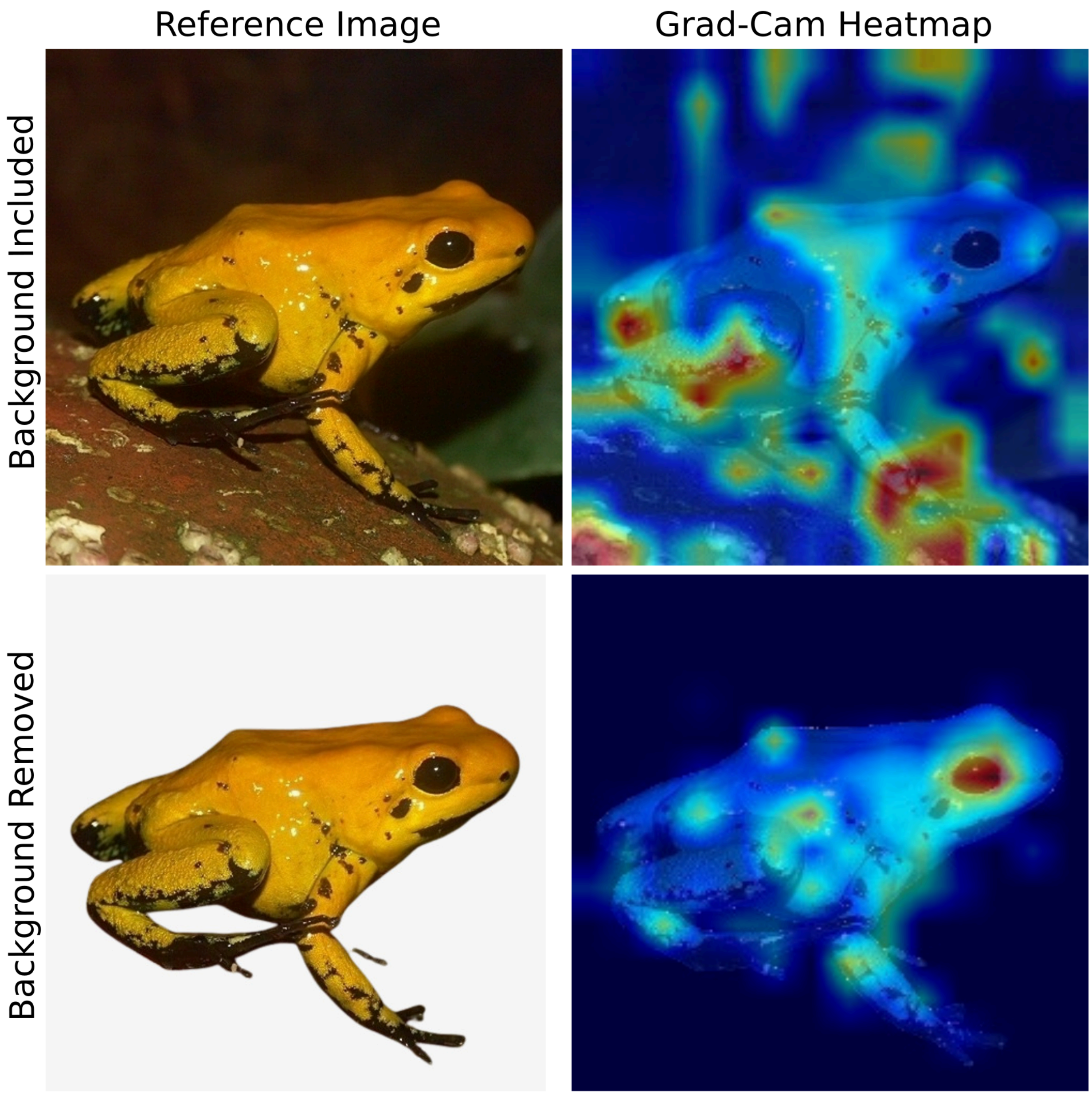

Figure 6: Effect of background on model attention during species classification. BioCLIP model attention maps for *Phyllobates terribilis* Myers, Daly & Malkin, 1978 classification task.

The top row shows the original image with the natural background intact, while the bottom row displays the same specimen with the background removed. Gradient-based heatmaps visualize regions that most strongly influence the model's species classification decision, with warmer colors indicating greater attention. Note how attention is more diffused across background habitat elements in the original image (potentially learning habitat associations), while it concentrates more specifically on specimen morphological features when the background is removed. For museum specimen imaging, this demonstrates the value of standardized backgrounds that direct model attention toward taxonomically diagnostic features rather than collection-specific environmental cues.. Image source: Hank Wallays, "Golden Poison Frog," 2005. Accessed via https://eol.org/media/8960654. CC BY-NC 3.0.

The prominence of background information may be adjusted by considering the focal plane of the image. Objects sitting at the focal plane (more accurately characterized as a curved surface) will be maximally focused, with increasing blur for objects further or nearer to the lens. By using smaller apertures, longer focal length lenses, and increasing the distance between the object and the lens, the photographer can increase the depth of field – the distance between the apparent loss of focus on either side of the focal plane. A narrow depth of field can be useful for blurring unwanted backgrounds such as underlying specimen labels or patterns, whereas a wider depth of field can be useful for specimens that are highly three-dimensional with parts near and far from the lens. However, setting the aperture much wider or much smaller than a particular lens' optimal F-stop will cause a loss of image clarity for optical reasons other than focus. In this case, another approach to increasing the depth of field is to use post-processing software to combine multiple stacked images with different focal planes to obtain a single image with all desired parts in maximal focus (Brecko et al., 2014).

**3.6 Lighting**

For biological specimen digitization, the primary objective should be faithful representation of specimen morphology and coloration. Lighting represents one of the most challenging aspects of imaging standardization, as optimal approaches vary significantly by specimen type and institutional resources.

Controlled artificial lighting (whether continuous or flash) provides the most consistent results by minimizing shadows, reflections, and color variations that can distort specimen features and complicate automated analysis. Flash photography offers the advantage of shorter shutter speeds, particularly valuable for macro photography where camera vibrations can blur images during longer exposures. However, both flash and continuous lighting can generate glare on glossy surfaces such as insect exoskeletons or wet plant tissues. Diffusion techniques such as softboxes, diffusion panels, multiple light sources, and light tents help scatter light evenly and reduce harsh reflections. When using flash, multiple units positioned to avoid direct reflection angles typically yield optimal results.

While controlled lighting is preferable, some imaging situations require natural light. When natural light is unavoidable, comprehensive metadata documentation becomes critical: record time of day, weather conditions, cardinal direction of light source, season, and any supplemental lighting used. This detailed documentation enables downstream researchers to account for lighting variability in their analyses.

Complete standardization across different camera systems is often impractical, particularly for multi-institutional projects. Metadata documentation provides the essential solution for managing this variability. Modern cameras automatically record critical settings in EXIF data, which should be preserved and linked to specimen records (see section 3.1). The goal should be consistency within individual projects while maximizing information content. When lighting conditions must vary due to specimen requirements or logistical constraints, comprehensive documentation enables downstream researchers to account for these differences in their analyses.

### 3.7 Resolution and Magnification

Image requirements for CV applications must balance model processing capabilities, storage constraints, and feature preservation needs (Hudson et al., 2015). The combination of resolution and magnification determines the effective pixel density – the number of pixels that represent a specific physical feature, determining how much detail is captured for analysis – on features or traits of interest. When imaging with a digital camera, this means selecting appropriate lens magnification and camera-to-subject distance to ensure that features of interest occupy sufficient pixels in the final image. For scanning biological imaging, an absolute minimum of 300 DPI is recommended for accuracy in measurements and detail preservation (Herler et al., 2008).

Appropriate magnification and resolution settings depend on the variation in specimen size and feature complexity. Microscopic organisms (e.g., mites, diatoms) require high magnification with high resolution to resolve fine structures, while larger specimens (e.g., plants, vertebrates) need less magnification to capture relevant traits. The goal is to ensure that key diagnostic features span sufficient pixels in an image for algorithm recognition. When image downsizing is necessary due to computational limitations, it is important to apply consistent parameters across the image dataset rather than mixing resolutions. It is also essential to maintain backups of original images and to document capture settings and any downsizing procedures in metadata to ensure reproducibility.

### 3.8 File Formats

During imaging, the raw version of the image should always be retained, however, other file formats are often required for image processing and storage. For scientific applications, lossless formats such as TIFF and PNG are preferable because they preserve every pixel of the original image. TIFF is the gold standard in biological imaging, offering high bit-depth storage and compatibility with scientific analysis software. The TIFF format ensures that all details remain intact, making it ideal for precise trait measurements. One of the most common causes of image degradation is the use of JPEG compression, which can irreversibly remove image data to reduce file size, leading to artifacts such as ringing, blocking, color bleeding, and color posterization in the image (X. Feng & Allebach, 2006; Tolstaya et al., 2010). Lost details cannot be restored to a JPEG image, posing a risk for biological studies where even minor distortions can lead to measurement inaccuracies. If images need to be compressed to reduce storage requirements, we recommend using PNG, which is lossless and is a standard format used in computer vision. If this is insufficient, state-of-the-art image compression tools such as AVIF (Barman & Martini, 2020) and JPEG-XL (Alakuijala et al., 2019) should be used, each of which supports both lossless and lossy quality options.

**3.9 Data Archiving and Storage**

Copies of datasets (images and metadata) should always be maintained in redundant locations to prevent loss of data (e.g., on a hard drive and in cloud storage). For organizational efficiency, maintain consistent folder hierarchies that separate raw images, processed derivatives, and metadata files. Storage and bandwidth requirements scale significantly with collection size. While small collections can be managed with standard external drives and cloud services, collections exceeding 10,000 specimens typically require network attached storage systems and systematic backup protocols following established data management standards (Wilkinson et al., 2016). Large-scale digitization projects benefit from early consultation with institutional IT professionals and data managers to develop appropriate infrastructure and backup strategies. Integration with existing institutional data management plans ensures long-term sustainability and compliance with funding agency requirements.

When images undergo processing, the alterations should be clearly documented in the imaging methods and metadata and applied consistently across all images in a dataset. This may apply to the process of downsizing images, as well as any color correction, distortion correction, or background removal. This information can be included in a file (e.g., a dataset card) (Gebru et al., 2021) or a general write-up of the data. This document would be published alongside the images and the metadata (CSV, JSON) described earlier, as it provides the context for the overall dataset (McMillan-Major et al., 2021). Additionally, the machine-readability of this information enhances the reusability of the dataset.

**3.10 Data Sharing**

After going through the steps to provide data of value to biologists and computer scientists, a final step is to enable further scientific insights by sharing the data. When archiving data for sharing, it is important to provide data on platforms discoverable by biologists (e.g., Zenodo, Figshare) and computer scientists (e.g., Hugging Face, Kaggle, OpenML, Dataverse) (NeurIPS, 2025). Maintaining records through these platforms improves the findability, accessibility, interoperability, and reusability of data for scientific discovery. When sharing data, it is important to select a license that the data collectors agree on to ensure attribution when desired. Best practice includes utilizing data sharing licenses with clear reuse policies (such as Creative Commons "CC BY" or "CC BY-SA") to remove ambiguity about how others may build upon the work. Furthermore, some of these platforms (notably Hugging Face) enable version control, which in combination with documentation can help end users to understand nuances between derivatives of the original images (e.g., batched images versus different methods for segmenting individual organisms from the batched images). The generation of persistent identifiers like DOIs by data sharing platforms is also an essential component of FAIR data principles, enabling proper citation and improving long-term discoverability (Wilkinson et al., 2016).

# 4. Implementation Framework and Future Standards Development

The ten considerations outlined above provide a conceptual framework for optimizing specimen imaging for computer vision applications. However, successful implementation requires translating these principles into actionable protocols. This section provides guidance for implementation while identifying critical areas where community-developed standards would enhance reproducibility and interoperability across research efforts.

### 4.1 Implementation Checklist

The following checklist summarizes key implementation steps, organized by generality of recommendations across versus within project:

Table 2: Fundamental Protocols (Standardized Across Projects) organized by consideration with development needs highlighted with an asterisk*.

| **Consideration** | **Protocol** |
|---|---|
| Metadata | Preserve link between images and unique specimen records (e.g., catalog number, unique ID) in metadata spreadsheet. |
| | Ensure redundancy in metadata capture to guard against data lost and enable critical insights into transcription errors (e.g., within spreadsheet and within image file). |
| | Maintain EXIF information where feasible, especially for camera and lens metadata. |
| | Record all methodological choices (e.g., equipment settings, lighting, post-processing) in metadata and README documentation. |
| File Formats * | Use lossless formats (TIFF or PNG preferred) for archival images, and maintain RAW files. *File name conventions - See Section 4.2 |
| Data Archiving and Storage | Backup data and maintain redundant storage (e.g., local and remote/cloud), with version control. |
| | Implement hierarchical file folders to separate data copies across processing steps |
| Data Sharing | Follow FAIR data principles and share data openly on platforms discoverable by both biologists and computer scientists (e.g., Zenodo, Figshare, Hugging Face, Kaggle) |

Table 3: Project-Specific Protocols (Standardized Within Projects) organized by consideration with development needs highlighted with an asterisk*.

| **Consideration** | **Protocol** |
|---|---|
| Specimen positioning | Define orientation protocols that minimize lens distortion and maximize reproducibility across images. |
| Size and Color Calibration * | Include standardized scale bars and color reference palettes in the same focal plane as the specimen, and avoid placement of these references in regions of lens distortion. *Standard placement - See Section 4.2 |
| Multiple Specimens | Arrange individuals with space between them to allow for optimized detection and cropping |
| | For cropped individuals, maintain clear linkages back to metadata tied to the original image |
| Background | Use uniform, non-reflective backgrounds that contrast appropriately with your specimens. For museum specimens, strive for uniformity of backgrounds, whereas for in-situ images, remember that the background may contain important habitat information. |
| Lighting | Choose either a consistent artificial lighting setup (e.g., LED panels or ring lights, opposing flashes) or establish consistency through documentation of natural light conditions when applicable (e.g., white balance, time of day, diffuser type). |
| Resolution and Magnification * | Use lenses with sufficient optical quality to resolve intended morphological features (e.g., use macro or enlarger lenses where appropriate) and test lens sharpness across the frame. *Pixel density - See Section 4.2 |

Equipment Selection Guidelines (Based on Specimen Characteristics):

Equipment selection should be validated through pilot testing with representative specimens from your collection to ensure that desired resolution, measurement accuracy, and workflow efficiency are possible before an extensive imaging campaign. Equipment choices are driven by specimen size, depth, shape, and measurement needs. These recommendations draw from current museum digitization practice and biological imaging literature. Other viable configurations may exist depending on local resources and specimen idiosyncrasies.

Table 4: Equipment Selection recommendations and considerations

| Specimen Type | Recommended Setup | Use Cases / Notes |
|---|---|---|
| **Flat, large specimens** (e.g., herbarium sheets, large leaves, pressed algae) | Flatbed scanners (e.g. Epson Expression 12000XL); DSLR/mirrorless on copy stand or conveyor belt | High resolution, minimal distortion. Scanners are ideal for resolution; cameras are preferred for speed (Tegelberg et al. 2014, Sweeney et al. 2018). |
| **Small to Medium, three-dimensional specimens** (e.g., insects, seeds, small mammals) | DSLR/mirrorless + macro lens + diffuse lighting | Enables precise morphology capture at small scale with adequate depth of field (Schmidt et al. 2012). |
| **Very Small, three-dimensional specimens** | Digital Microscope with image stacking capability | Captures high resolution images, but is labor intensive. |
| **Pinned specimens in drawers or high volume collections** | Large-format rail systems, medium-format stitched imaging, or tray/drawer-level DSLRs | Captures multiple specimens; additional processing and metadata needed (Blagoderov et al. 2012). **May not be appropriate for all specimen sizes, so consider pixel density carefully.** |
| **Glass microscope slides** | Whole-slide scanners | Low Cost, open source options available (Lu et al. 2018). |
| *Note: These configurations are not exhaustive. Many institutions customize equipment combinations based on budget, existing infrastructure, and institutional needs. Open-source or DIY alternatives may offer cost-effective solutions for many tasks.* | | |

**4.2 Call for Community Action to Develop Imaging Standards**

The evidence presented in this review demonstrates both the potential benefits of standardized imaging protocols and the current lack of community consensus on implementation details. Several critical areas require coordinated community development to maximize interoperability and effectiveness across institutions. Effective pixel density standards represent the most urgent need, as minimum requirements for different trait extraction tasks across taxonomic groups remain largely undefined. Pilot studies are needed to systematically assess minimum imaging requirements for various computer vision applications, particularly given the significant variation in specimen size and morphological complexity across taxa. Similarly, developing consistent filename conventions aligned with established metadata standards such as Darwin or Audiovisual Core would facilitate large-scale dataset integration, moving beyond current institutional-specific approaches that create barriers to cross-institutional research.

Community guidelines for color and scale bar positioning would enhance both measurement accuracy and automated processing capabilities, while coordinated standards for imaging practices, data structure, and metadata will facilitate integration across digitization efforts. We recommend collaborative steps including: developing naming conventions for image files aligned with metadata standards, conducting pilot studies to determine minimum pixel density requirements across taxonomic groups and traits, and curating reference datasets that illustrate best practices for new digitization programs. These efforts should be coordinated through existing professional societies and digitization networks to ensure broad adoption and community buy-in.

The transition from traditional digitization to computer vision-optimized imaging represents a significant opportunity to enhance the scientific value of biological collections. While implementing these recommendations may require additional investment in equipment, training, and time, the resulting datasets will support more reliable automated analysis, enabling novel interdisciplinary research at previously impossible scales. We encourage collection managers, taxonomists, and biodiversity informaticians to view these guidelines as adaptable principles that can be tailored to specific research questions, taxonomic groups, and institutional capabilities. The key to successful adaptation lies in maintaining thorough documentation of metadata and methodology choices made for each consideration. When researchers document decisions about lighting, positioning, background selection, and other variables discussed here, they create datasets that remain useful even as imaging techniques and analytical approaches evolve.

Success will depend on sustained collaboration between collection managers, taxonomists, computer scientists, and informaticians to create standards that serve both immediate practical needs and long-term scientific goals. This comprehensive documentation ensures transparency, reproducibility, accessibility, reusability, and the ability to identify potential sources of bias or error in downstream analyses. As computer vision technology advances, this detailed record-keeping will become increasingly valuable, helping to realize the full scientific potential of digitized biological collections.

The collaborative standards development process outlined above provides a pathway for the community to build upon this foundation, ensuring that biological collections can fully participate in the computer vision revolution while maintaining the scientific rigor that makes these collections invaluable for understanding biodiversity.

## Acknowledgments

This work was supported by both the Imageomics Institute and the AI and Biodiversity Change (ABC) Global Center. The Imageomics Institute is funded by the US National Science Foundation's Harnessing the Data Revolution (HDR) program under Award #2118240 (Imageomics: A New Frontier of Biological Information Powered by Knowledge-Guided Machine Learning). The ABC Global Climate Center is funded by the US National Science Foundation under Award #2330423 and the Natural Sciences and Engineering Research Council of Canada under Award #585136. SR and AE were additionally supported by NSF award #242918 (EPSCOR Research Fellows: NSF: Advancing National Ecological Observatory Network-Enabled Science and Workforce Development at the University of Maine with Artificial Intelligence) and by Hatch project Award #MEO-022425 from the US Department of Agriculture's National Institute of Food and Agriculture. This material is based in part upon work supported by the National Ecological Observatory Network (NEON), a program sponsored by the U.S. National Science Foundation (NSF) and operated under cooperative agreement by Battelle